# ABCAS: Adaptive Bound Control of spectral norm as Automatic Stabilizer

Shota HIROSE, Shiori MAKI,
Naoki WADA, Jiro KATTO
School of Fundamental Science and Engineering
Faculty of Science and Engineering, Waseda University
Tokyo, Japan
syouta.hrs@akane.waseda.jp

Heming SUN
Waseda Research Institute for Science and Engineering
Waseda University, Tokyo, Japan
and JST, PRESTO, Saitama, Japan
hemingsun@aoni.waseda.jp

*Abstract—* Spectral Normalization is one of the best methods for stabilizing the training of Generative Adversarial Network. Spectral Normalization limits the gradient of discriminator between the distribution between real data and fake data. However, even with this normalization, GAN's training sometimes fails. In this paper, we reveal that more severe restriction is sometimes needed depending on the training dataset, then we propose a novel stabilizer which offers an adaptive normalization method, called ABCAS. Our method decides discriminator's Lipschitz constant adaptively, by checking the distance of distributions of real and fake data. Our method improves the stability of the training of Generative Adversarial Network and achieved better Fréchet Inception Distance score of generated images. We also investigated suitable spectral norm for three datasets. We show the result as an ablation study.

## I. Introduction

Generative Adversarial Network (GAN) [1], is one of the successful neural networks. GAN learns the distributions of data by training a discriminator and a generator at the same time, and then generates fake data by the generator. It can synthesize images with high quality.

However, training of GAN is often unstable. To solve this problem, various approaches have been proposed. For example, WGAN [2] limits the Lipschitz constant of the discriminator by applying Weight Clipping. It stabilizes GAN's training because this regularization limits the upper bound of the norm of the gradient which the generator receives. However, Weight Clipping is not a safe way, and the performance depends on clipping values. If the clipping value is too small, it will be difficult to train the discriminator, and if clipping value is too large, mode collapse will occur. As alternative way to restrict the Lipschitz constant, Spectral Normalization [3] is proposed. It normalizes spectral norm, which is the upper bound of Lipschitz constant. This method drastically stabilizes the training of GAN with justifiable computational cost. However, even with this normalization, the training of GAN sometimes fails.

In this paper, we therefore propose a new method, called ABCAS. It controls the distance of real and fake distributions by limiting the spectral norm of the discriminator adaptively. We show the suitable norms are different depending on datasets. We then confirm that our method works well even in such situations.

## II. Related work

In Generative Adversarial Network, a generator tries to deceive a discriminator by generating images which look real, while the discriminator tries to classify input images to real or fake. GAN's training is often unstable and mode collapse often occurs. To mitigate such problems, various approaches have been proposed. For example, definition of loss functions has been modified [2, 4, 5].

Instability in GAN's training often appears with limited data. Many researchers have been looking for good methods for mitigating this problem. To our knowledge, it is because the discriminator overfits the training dataset due to data insufficiency, and there are two ways to avoid the instability.

First is Data Augmentation, that can virtually mitigate the insufficiency and solve the overfitting. For example, Adaptive Discriminator Augmentation [6] applies the adaptive strategy for data augmentation by detecting the overfitting of the discriminator and controls the strength of Data Augmentation. Adaptive Pseudo Augmentation [7] is another method, which adds "real" label to generated data when overfitting occurs. The method improves the stability of GAN's training.

Second is model regularization. Overfitting is also caused by overparameterization. Therefore, model regularization can mitigate the overfitting. For example, Weight Clipping [2], Gradient Penalty [8], Spectral Normalization [3], and Label Noise [9] have been proposed.

Wasserstein GAN (WGAN) [2] minimizes Wasserstein distance between real data and fake data. To let the discriminator return approximation of Wasserstein distance, weights of the discriminator are clipped and Lipschitz constant is regularized.

Gradient Penalty [8] regularizes the Lipschitz constant by regularizing the gradient between real data and fake data. This method was proposed to improve WGAN.

Spectral Normalization [3] also limits the Lipschitz constant by limiting the spectral norm of the discriminator because the Lipschitz constant is bounded by the product of the spectral norms of weight matrices corresponding to convolutional filtering of each layer. In the algorithm of Spectral Normalization, the spectral norm of weight matrix is calculated by using power iteration method [10], that contributes to faster computation than singular value decomposition.

SAGAN [11] and BigGAN [12] also uses Spectral Normalization. In addition, these GANs use Self-Attention [13]. It improves the quality of generated images. However, it needs huge computation and memory cost.

Spectral Normalization performs well to stabilize GAN's training and has been utilized by many subsequent GANs. However, Spectral Normalization still suffers from instability depending on utilized datasets.

In this paper, we therefore show an adaptive normalization method by extending Spectral Normalization. Our method can stabilize GAN's training better than the original Spectral Normalization and other existing methods.

## III. PROPOSED APPROACH

To limit the Lipschitz constant more strictly than the Spectral Normalization, we propose a new method, Adaptive Bound Control of Spectral Normalization as Automatic Stabilizer (ABCAS). For this purpose, we introduce two parameters, $r$ and $beta$. $r$ is a variable to control magnitude of the normalized spectral norm of the discriminator by multiplying $m = 0.9^r$. $beta$ is a constant to limit the parameter $r$.

Our method aims to limit the distance between the distribution of real data and fake data less than constant value, $beta$, in ABCAS. Spectral Normalization limits the spectral norm of the discriminator, however, the output value is not limited to specific range because the maximum norm of the critic depends on the norm of input images.

We show the algorithm of ABCAS in Figure 1. The essential point is adaptive control of the spectral norm by checking the distance between real data and fake data and updating $r$ adaptively. If the distance between real data and fake data is large, $r$ becomes large and the spectral norm of the discriminator does smaller. This leads smaller distance between real data and fake data. In ABCAS, $beta$ is a hyper-parameter. If $beta$ is small, ABCAS limits the spectral norm more severely. The result of our experiment shows that suitable $beta$ is 4.

```
m,r,dm,counter ←0 #initialization
alpha←0.9999 #for running average
while training:
    counter←counter+1
    sample z from N(0,1)
    x_real←loadimage(dataset)
    x_fake←G(z)
    for each matrix W of discriminator:
        compute the spectral norm σ(W)
    m = 0.9^r
    W' = m* W / σ(W)  #ABCAS
    C_real ←D(x_real) using W'
    C_fake ←D(x_fake) using W'
    if value of counter is odd:
        dist← max(C_real) - min(C_fake)
        dm ← alpha * dm + (1-alpha) * dist
        clbr_dm← dm/beta
    r ← max(0, clbr_dm/(1-clbr_dm))
    D_loss←loss_function(x_real, x_fake)
    G_loss←loss_function(x_fake)
    If value of counter is odd:
        update Discriminator
    else: #value of counter is even:
        update Generator
```

Fig. 1. Pseudocode of ABCAS

## IV. EXPERIMENTS

### A. Network architecture

We use a network based on SNGAN[3]. We scale up the discriminator and the generator to generate images with 256x256 pixels. We apply ABCAS for all convolution layers of the discriminator and we don't use Batch Normalization[14]. For the generator, we remove Batch Normalization and add Layer Normalization[15] after the second last convolution layer. In the paper[16], they show Layer Normalization can stabilize the training well, so we use Layer Normalization. In addition, we add Pixel Normalization[17] in the first layer of the generator to normalize the latent input vector. For the training, we use non-saturating GAN loss [1]. We use RAdam[18] optimizer with $β_1 = 0$, $β_2 = 0.999$. The batch size is 16. We use TTUR[19] technique to accelerate the training of GAN. The learning rate of the discriminator is 0.0005 and the learning rate of the generator is 0.0001. To make sure that our method works well for various datasets, we use FFHQ[20] dataset, AFHQ-Cat[21], and FFHQ-10k(a subset of FFHQ, it consists of 10,000 images) as training dataset. AFHQ-Cat dataset consists of about 5,165 images of faces of cats. FFHQ dataset consists of about 70,000 images of faces of people. We created FFHQ-10k dataset by choosing first 10,000 images of FFHQ dataset. For all datasets, we resized them to 256x256 resolution, using MATLAB.

The network details which we used are shown in Table I and Table II.

TABLE I. NETWORK OF GENERATOR WE USE

| |
| --- |
| z∈R^140 ~N(0,1) |
| Pixel Normalization(z) #normalize z channel-wise |
| 4x4 stride=1, convtranspose, 384x4x4, LReLU(0.2) |
| 4x4 stride=2, convtranspose, 192x8x8, LReLU(0.2) |
| 4x4 stride=2, convtranspose, 96x16x16, LReLU(0.2) |
| 4x4 stride=2, convtranspose, 96x32x32, LReLU(0.2) |
| 4x4 stride=2, convtranspose, 48x64x64, LReLU(0.2) |
| 4x4 stride=2, convtranspose, 24x128x128, Layer Normalization, ReLU |
| 4x4 stride=2, convtranspose, 3x256x256, Tanh |

TABLE II. NETWORK OF DISCRIMINATOR WE USE

| |
| --- |
| RGB image x∈R^(3x256x256) |
| 4x4, stride=2, ABCASConv 24x128x128, ReLU |
| 4x4, stride=2, ABCASConv 48x64x64, ReLU |
| 4x4, stride=2, ABCASConv 96x32x32, ReLU |
| 4x4, stride=2, ABCASConv 96x16x16, ReLU |
| 4x4, stride=2, ABCASConv 192x8x8, ReLU |
| 4x4, stride=2, ABCASConv 384x4x4, ReLU |
| 4x4, stride=1, ABCASConv 384x1x1 |

### B. Comparisons of Fréchet Inception Distance scores

To verify that our adaptive method works better than original Spectral Normalization, we compare with Spectral Normalization with fixed $m$ values. With ABCAS algorithm, we can adjust the spectral norm of each layer by changing $m$ in the algorithm, that is represented by $m = ABCAS(beta)$. Since the original Spectral Normalization is corresponding to the case $m = 1$, we try cases $m = 0.5, 0.6, 0.7, 0.8$ and $0.9$, that bring smaller spectral norms than the original Spectral Normalization. We show experimental results with parameter settings in Table III. In Table III, "Spectral Norm" represents fixed $m$ in the Spectral Normalization and adaptive $m$ in the ABCAS algorithm.

TABLE III. COMPARISON OF GANs AND THE BEST FID

| Dataset | Spectral Norm | Best FID |
|---|---|---|
| AFHQ (Cat) | 0.5 | 47.68 |
|  | 0.6 | 35.30 |
|  | 0.7 | 31.23 |
|  | 0.8 | 28.50 |
|  | 0.9 | 26.90 |
|  | 1.0 | 33.25 |
|  | ABCAS(beta=1) | 33.99 |
|  | ABCAS(beta=4) | **24.99** |
| FFHQ | 0.5 | 53.54 |
|  | 0.6 | 50.96 |
|  | 0.7 | 40.96 |
|  | 0.8 | 34.88 |
|  | 0.9 | 35.27 |
|  | 1 | 34.33 |
|  | ABCAS(beta=1) | 35.30 |
|  | ABCAS(beta=4) | **34.20** |
| FFHQ-10k | 0.5 | 83.67 |
|  | 0.6 | 90.81 |
|  | 0.7 | 70.30 |
|  | 0.8 | 65.27 |
|  | 0.9 | 59.80 |
|  | 1 | 68.59 |
|  | ABCAS(beta=1) | 56.12 |
|  | ABCAS(beta=4) | **55.48** |

For FFHQ dataset, we test each setting for 200 epochs, and measure Fréchet inception distance [19] (FID) every epoch. For AFHQ-Cat and FFHQ-10k dataset, we test each setting for 1000 epochs, and measure FID every 5 epochs. We use GeForce RTX 3070 as GPU. We show relationship between elapsed epoch and FID in the training in Fig 2, where (a) is for AFHQ-Cat dataset, (b) is for FFHQ dataset, and (c) is for FFHQ-10k dataset, respectively. Horizontal axis denotes elapsed epochs on a linear scale, and vertical axis denotes FIDs on a logarithmic scale. Legends of each figure correspond to the settings in Table IV.

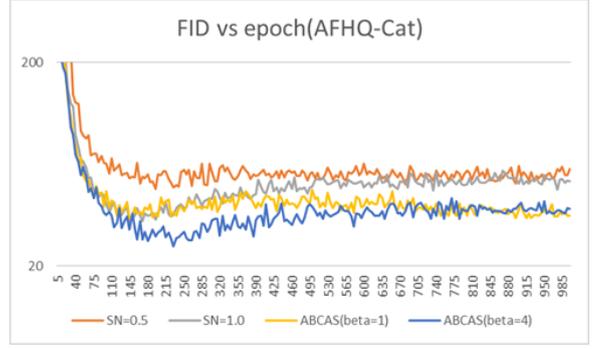

(a) AFHQ-Cat

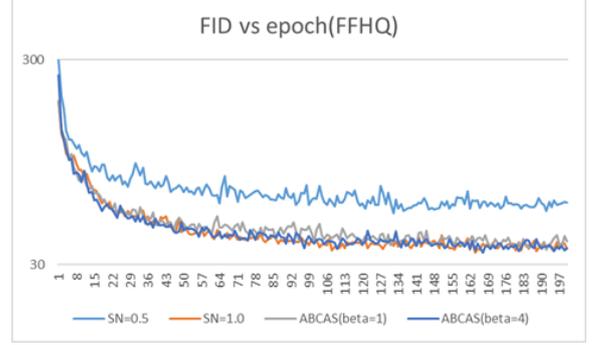

(b) FFHQ

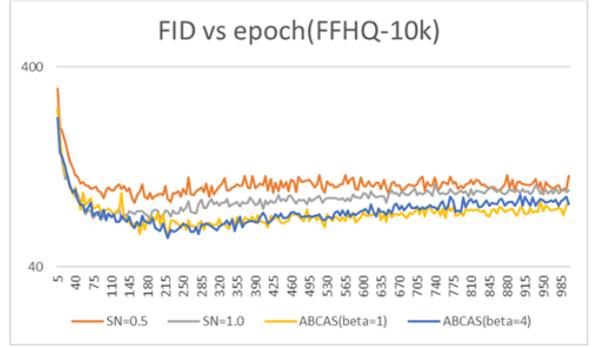

(c) FFHQ-10k

Fig. 2. Comparisons of elapsed epoch vs FIDs

From these results, we can recognize that the original Spectral Normalization ($m$=1) is not the best, and optimal $m$ changes according to datasets. We notice that ABCAS(beta=1) performs better than the original Spectral Normalization for FFHQ-10k dataset, and ABCAS(beta=4) performs the best for all three datasets.

Next, we show how the parameter $r$ in ABCAS(beta=4) changes in training phase. Fig.3 shows that $r$ rises rapidly and fall slightly in early training. After that, $r$ rises gradually, and $m$ ($= 0.9^r$) becomes smaller. It is also confirmed that final $r$ is different, depending on utilized datasets. We think $r$ can be higher when dataset is small, considering that the result of final $r$ is the largest in AFHQ-Cat (5k images), followed in order by FFHQ-10k (10k images) and FFHQ (70k images).

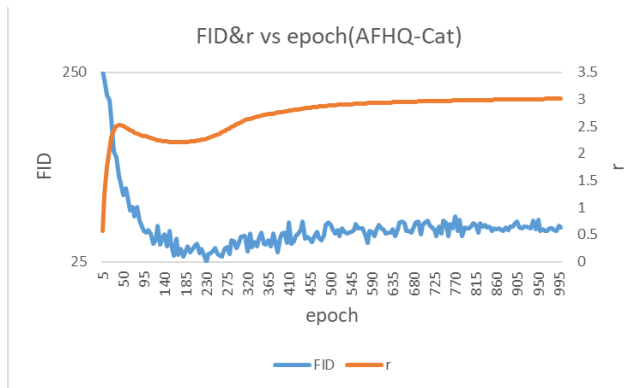

(a) AFHQ-Cat

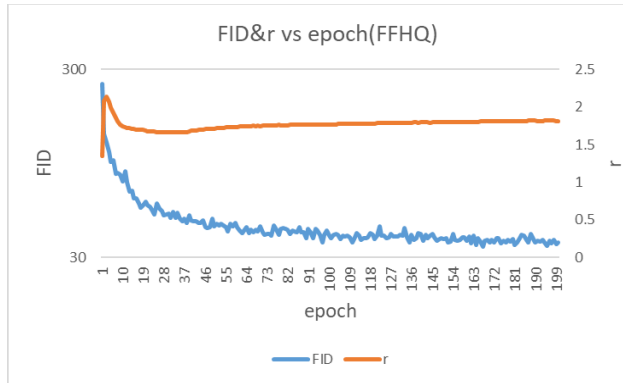

(b) FFHQ

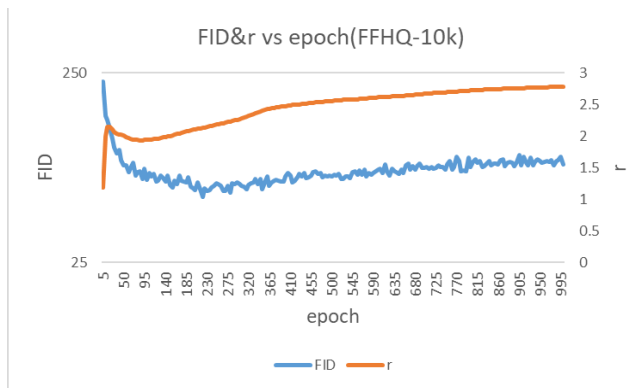

(c) FFHQ-10k

Fig 3. Comparisons of FIDs and parameter *r* per epoch

*C. Subjective comparisons of generated images*

To compare perceptual differences of the results, we show generated images with AFHQ-Cat dataset. In Fig.4, we show generated images of the cases that $m = 0.5$, $m = 1$ and ABCAS($beta = 4$).

In Fig.4(a), when $m$ is too small, FID is worse than the original Spectral Normalization ($m = 1$). From the subjective viewpoint, generated images are visually worse than other two cases. Generated images are blurrier, and the cat in upper right has unnatural eyes and strange face contour. Therefore, we can conclude that $m = 0.5$ is not a good choice.

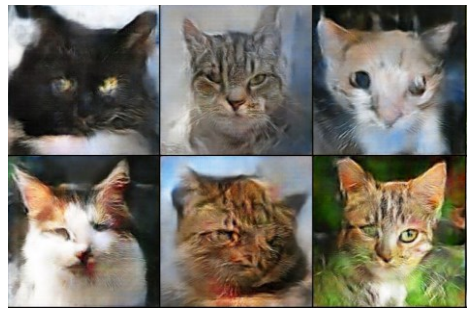

(a) Constant SN, $m = 0.5$ (195 epochs, FID=47.68)

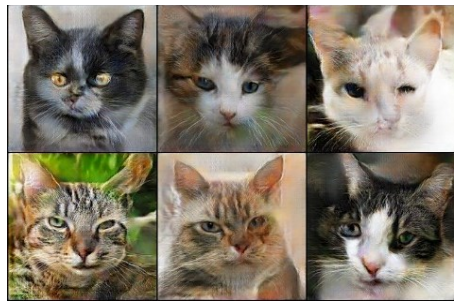

(b) Constant SN, $m = 1.0$ (145 epochs, FID=33.25)

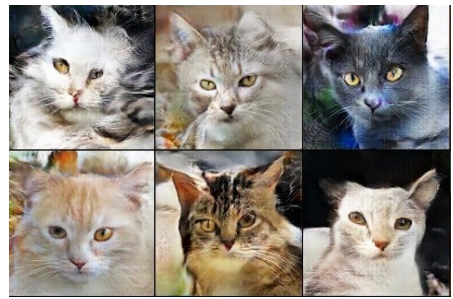

(c) ABCAS($beta = 4$) (230 epochs, FID=24.99)

Fig. 4: Subjective comparisons of generated images.

The original Spectral Normalization in Fig.4(b) and ABCAS ($beta = 4$) in Fig. 4(c) clearly provide better generated images then Fig. 4(a) even if we used very simple GAN. However, the cats in upper left and upper right of Fig.4(b) have unnatural noses, and the cats in lower left and lower right have strange contour. On the other hand, all cats in Fig.4(c) have natural eyes and noses. Therefore, we can conclude that the overall subjective quality of ABCAS ($beta = 4$) is higher than Spectral Normalization with fixed $m$.

V. CONCLUSIONS

We proposed ABCAS, which controls spectral norms in an adaptive manner to image datasets. Although our proposed algorithm is very simple, experimental results verified that GAN's training becomes more stable. In addition, our experiment results show that preferable spectral norms are different, depending on datasets which are used in training phase. We continue to look for better metrics for measuring the distance of real data and fake data for improving our method.